\documentclass[12pt]{elsarticle}
\usepackage[margin=1in]{geometry}

\usepackage[utf8]{inputenc} 
\usepackage[T1]{fontenc}    
\usepackage{hyperref}       
\usepackage{url}            
\usepackage{booktabs}       
\usepackage{amsfonts}       
\usepackage{nicefrac}       
\usepackage{microtype}      
\usepackage{lipsum}
\usepackage{graphicx}
\RequirePackage{amsmath}
\usepackage{tabularx}
\usepackage{colortbl}
\usepackage[table]{xcolor}
\usepackage{algorithm,algpseudocode}
\usepackage{graphicx,subcaption}

\begin{document}
\begin{frontmatter}

\journal{Journal of Computational Physics}
\title{Enabling Probabilistic Learning on Manifolds \\
 through Double Diffusion Maps}

\author[1]{Dimitris G. Giovanis\corref{cor1}}\ead{dgiovan1@jhu.edu}
\author[2]{Nikolaos Evangelou}\ead{nevange2@jh.edu}
\author[2,3]{Ioannis G. Kevrekidis\corref{cor1}}\ead{yannisk@jhu.edu}
\author[4,5]{Roger  G. Ghanem\corref{cor1}}\ead{ghanem@usc.edu}
\address[1]{Dept. of Civil \& Systems Engineering, Johns Hopkins University, USA}
\address[2]{Dept. of Chemical and Biomolecular Engineering, Johns Hopkins University, USA}
\address[3]{Dept. of Applied Mathematics and Statistics, Johns Hopkins University, USA}
\address[4]{Dept. of Civil and Environmental Engineering, University of Southern California, USA}
\address[5]{Dept. of Aerospace and Mechanical Engineering, University of Southern California, USA}

\cortext[cor1]{Corresponding authors.} 

\begin{abstract}
We present a generative learning framework for probabilistic sampling based on an extension of the Probabilistic Learning on Manifolds (PLoM) approach, which is designed to generate statistically consistent realizations of a random vector in a finite-dimensional Euclidean space, informed by a limited (yet representative) set of observations. In its original form, PLoM constructs a reduced-order probabilistic model by combining three main components: (a) kernel density estimation to approximate the underlying probability measure, (b) Diffusion Maps to uncover the intrinsic low-dimensional manifold structure, and (c) a reduced-order Itô Stochastic Differential Equation (ISDE) to sample from the learned distribution. A key challenge arises, however, when the number of available data points $N$ is small and the dimensionality of the diffusion-map basis approaches $N$, resulting in overfitting and loss of generalization. To overcome this limitation, we propose an enabling extension that implements a synthesis of Double Diffusion Maps—a technique capable of capturing multiscale geometric features of the data—with Geometric Harmonics (GH), a nonparametric reconstruction method that allows smooth  nonlinear interpolation in high-dimensional \textit{ambient} spaces. This approach enables us to solve a full-order ISDE directly in the latent space, preserving the full dynamical complexity of the system, while leveraging its reduced geometric representation. The effectiveness and robustness of the proposed method are illustrated through two numerical studies: one based on data generated from two-dimensional Hermite polynomial functions and another based on high-fidelity simulations of a detonation wave in a reactive flow.

\end{abstract}

\end{frontmatter}

\section{Introduction}
\noindent
Generative modeling has emerged as a foundational paradigm in computational science, offering a data-driven framework for simulating complex systems and exploring latent patterns in physical phenomena \cite{crabtree2024micro, gao2025generative, gao2024generative, crabtree2025generative, giovanis2025generative}. These models are designed to learn high-dimensional probability distributions from limited or noisy data and to synthesize new, high-fidelity samples that are \textit{statistically consistent} with the training distribution. In doing so, generative models capture not only the central tendencies of a dataset, but also its underlying structure, variability, and uncertainty—properties that are essential for robust prediction, inference, and scientific discovery. By modeling the full distribution rather than only point estimates, generative models provide a principled way to handle epistemic and aleatoric uncertainty [REF]. This capability is crucial in applications where data is scarce, expensive to collect, or governed by partially known physical laws. In such settings, generative models can interpolate or extrapolate plausible data realizations, thereby enabling downstream tasks such as data augmentation, surrogate modeling, anomaly detection, and sensitivity analysis. Moreover, their ability to integrate prior knowledge, enforce physical constraints, or embed geometric structure (e.g., manifolds) into the learning process makes them particularly well-suited for modeling scientific and engineering systems.

Beyond synthetic data generation, generative models play a critical role in uncertainty quantification, system identification, and model updating in digital twin frameworks. They enable the assimilation of observational data into computational models, support adaptive experimentation by suggesting informative queries, and facilitate the simulation of counterfactual scenarios. As a result, they have become indispensable tools across a wide range of disciplines, including molecular biology~\cite{jumper2021highly}, neuroscience~\cite{wang2023applications}, physics~\cite{de2017learning}, materials science~\cite{dan2020generative}, and climate science~\cite{besombes2021producing}. As the field continues to evolve, the integration of generative modeling with domain-specific priors, geometric learning and probabilistic reasoning promises to unlock new frontiers in scientific understanding and discovery.

At the core of many generative modeling approaches is the notion of learning a map between samples of specified probability distributions. This enables the transformation of samples from a known reference distribution into those of a target distribution defined implicitly by observed data. State-of-the-art techniques utilize deep neural networks (DNNs) to parameterize this transformation, typically through a stochastic process—such as an Itô stochastic differential equation (ISDE)—whose solution evolves from simple noise to structured observations.  Modern generative models fall into three broad categories: likelihood-based methods (e.g., variational autoencoders~\cite{kingma2013auto, rezende2014stochastic}, autoregressive models~\cite{larochelle2011neural}, and normalizing flows~\cite{papamakarios2021normalizing}), implicit models (e.g., generative adversarial networks~\cite{goodfellow2020generative, mirza2014conditional}), and score-based diffusion models~\cite{song2020score}. The latter class has shown remarkable success through the use of stochastic forward processes that corrupt data with noise, followed by a learned reverse process to recover realistic samples. In particular, score-based generative models (SGMs) estimate the gradient of the log-density (i.e., the \emph{score function}) to guide the generative dynamics through reverse-time SDEs~\cite{song2019generative, ho2020denoising}. These methods have demonstrated state-of-the-art performance in image~\cite{song2020improved}, audio~\cite{kong2020diffwave}, and graph generation~\cite{niu2020permutation}, and protein chemistry \cite{watson2022broadly}, among others. Despite their empirical success, diffusion-based generative models face several limitations when applied to structured scientific datasets. Specifically, they often rely on unconstrained Euclidean embeddings, which can obscure the low-dimensional, nonlinear structures—i.e., manifolds—on which many physical processes evolve. As such, these models may require large amounts of data or extensive training to accurately learn the geometry of the underlying distribution. 

Probabilistic Learning on Manifolds (PLoM) \cite{soize2016data, soize2019entropy, soize2021probabilistic, soize2022probabilistic} addresses these limitations by explicitly incorporating manifold structure into the learning process. Manifold learning is based on the assumption that high-dimensional data typically lie on a lower-dimensional manifold embedded within the high-dimensional space \cite{fefferman2016testing}. Examples of manifold learning methods include Isomap \cite{balasubramanian2002isomap}, local linear embedding (LLE) \cite{roweis2000nonlinear}, and Diffusion Maps  \cite{coifman2006diffusion}. PLoM approximates the data distribution by combining kernel density estimation with Diffusion Maps for nonlinear dimensionality reduction and an ISDE for sample generation on the manifold. However, traditional PLoM formulations are limited by their reliance on reduced-order dynamics and the lack of a principled inverse map from the latent manifold back to the ambient data space (what is called \textit{lifting} [REF]). To overcome these challenges, we introduce an extended PLoM framework that incorporates Double Diffusion Maps~\cite{evangelou2023double} and Geometric Harmonics (GH)~\cite{coifman2006geometric}, enabling the solution of a full-order ISDE in the reduced identified latent space and smooth lifting to the full data domain. This hybrid approach retains the geometric expressiveness of manifold learning while improving sample quality and generalization in low-data regimes.

In this work, we present a generative learning framework for sampling from probability distributions supported on low-dimensional manifolds embedded in high-dimensional ambient spaces.
More specifically, we introduce an extended PLoM framework that integrates Double Diffusion Maps~\cite{evangelou2023double} and Geometric Harmonics (GH)~\cite{coifman2006geometric}, enabling the solution of a full-order ISDE in the identified latent space and smooth lifting to the full data domain. This hybrid approach retains the geometric expressiveness of the original PLoM, while improving sample quality and generalization in low-data regimes. While Double Diffusion Maps was originally developed for constructing reduced-order models, here we employ them to discover a low-dimensional latent space via a first application of Diffusion Maps, followed by a second application on the latent coordinates. This two-step embedding allows for the recovery—or \textit{lifting}—of new samples back to the high-dimensional ambient space using the Nystr\"om extension~\cite{coifman2006geometric} and GH-based interpolation, ensuring smooth and consistent reconstructions.

The remainder of this paper is organized as follows. Section~\ref{sec:plom} reviews the original PLoM methodology and its mathematical underpinnings, while Section~\ref{sec:challenges} outlines its limitations and the motivation for the proposed extension. Section~\ref{sec:ddmaps} introduces Double Diffusion Maps and their role in the generative process. The proposed framework is detailed in Section~\ref{sec:method}, followed by numerical experiments and results in Section~\ref{sec:examples}. Finally, conclusions and directions for future work are discussed in Section~\ref{sec:conclusions}.

\section{Probabilistic Learning on Manifolds (PLoM)}
\label{sec:plom}

\noindent
Probabilistic Learning on Manifolds (PLoM) \cite{soize2016data} is designed to generate new realizations of random vectors in $\mathbb{R}^n$ whose probability distribution is concentrated on an unknown, low-dimensional manifold, embedded in a higher-dimensional space. Let $\textbf{x} = (x_1, \dots, x_n) \in \mathbb{R}^n$ denote a generic point in Euclidean space, and let $\text{d}\textbf{x} = \text{d}x_1 \cdots \text{d}x_n$ be the associated Lebesgue measure. Consider a dataset of $N$ $\mathbb{R}^n$-valued samples denoted by $\{\textbf{x}^1, \dots, \textbf{x}^N\}$. Moreover, let $\mathbf{X} = (X_1, \dots, X_n)$ be an $\mathbb{R}^n$-valued random vector defined on some probability space, and having some arbitrary probability density function (pdf) $p_{\textbf{X}}$ supported on a subset $S_n \subset \mathbb{R}^n$. The available dataset consists of $N$ independent realizations of $\mathbf{X}$, denoted as $\textbf{x}^{d,1}, \dots, \textbf{x}^{d,N} \in \mathbb{R}^n$. This dataset can be represented in matrix form as $[x_d] \in \mathbb{M}_{n,N}$, where $[x_d]_{kj} = x_k^{d,j}$, and construed as a sample of random matrix $[\mathbf{X}]$ with values in $\mathbb{M}_{n,N}$  The objective of PLoM to devise a methodology to generate samples from the  unknown probability distribution of $[\mathbf{X}]$, which is assumed to concentrate on an unknown subset $S_n \subset \mathbb{R}^n$, based solely on the (assumed representative) single realization $[x_d]$.  The PLoM sampling framework involves the following core steps, discussed next.

\subsection*{Data Preprocessing using Principal Component Analysis}

\noindent
The first step of PLoM involves preprocessing the dataset for numerical stability and statistical analysis. Given $N$ realizations of a random vector $\mathbf{X} \in \mathbb{R}^n$, organized in $[x_d^\text{uns}] \in \mathbb{M}_{n,N}$, we apply min-max normalization to obtain $[x_d] \in \mathbb{M}_{n,N}$:

\begin{equation}
[x_d]_{kj} = \frac{[x_d^\text{uns}]_{kj} - \min_{j'} [x_d^\text{uns}]_{kj'}}{\max_{j'} [x_d^\text{uns}]_{kj'} - \min_{j'} [x_d^\text{uns}]_{kj'}} + \varepsilon_s,
\end{equation}

\noindent
where $\varepsilon_s$ prevents division by zero. This normalization ensures consistent scaling across dimensions, essential for kernel-based operators. We then compute the empirical mean $\mathbf{m}$ and covariance $[\mathbf{c}]$:

\begin{align}
\mathbf{m} &= \frac{1}{N} \sum_{j=1}^N \textbf{x}^{d,j}, \quad
[\mathbf{c}] = \frac{1}{N - 1} \sum_{j=1}^N (\textbf{x}^{d,j} - \mathbf{m})(\textbf{x}^{d,j} - \mathbf{m})^T.
\end{align}

\noindent
Principal component analysis (PCA) is then performed by solving the eigenproblem:

\[
[\mathbf c] \boldsymbol{\phi}_k = \mu_k \boldsymbol{\phi}_k, \quad \boldsymbol\phi_k^T\boldsymbol\phi_k = 1, \quad k = 1, \dots, n,
\]

\noindent
Retaining the leading $\nu$ eigenvectors, we define $\boldsymbol\Phi = [\boldsymbol{\phi}_1, \ldots, \boldsymbol{\phi}_\nu] \in \mathbb{M}_{n,\nu}$. PCA projects the data onto this reduced basis:

\begin{equation}\label{eq:pca}
[\mathbf{X}] = [\underline{x}] + [\boldsymbol{\Phi}] [\mu]^{1/2} [\mathbf{H}],
\end{equation}

\noindent
where $[\underline{x}]$ replicates $\mathbf{m}$ across columns, $[\mu]$ is a diagonal matrix of eigenvalues, and $[\mathbf{H}] \in \mathbb{M}_{\nu,N}$ contains the reduced coordinates:

\begin{equation}
[\eta_d] = [\mu]^{-1/2} [\boldsymbol{\Phi}]^\top ([x_d] - [\underline{x}]).
\end{equation}

\noindent
This transformation reduces dimensionality from $n$ to $\nu$ and decorrelates the data, enabling efficient estimation of the latent probability structure and manifold geometry in the subsequent stages of PLoM.

\subsection*{Nonparametric Estimation of the Probability Density Function}

\noindent
After transforming the data into a reduced and normalized space via PCA, the next step is to estimate the underlying pdf of the random vector $\mathbf{H}$. Since its true pdf is unknown and no parametric form is assumed, a nonparametric estimation technique is used. The original PLoM adopts a kernel density estimation (KDE) approach, which constructs the pdf as a superposition of localized kernels placed at each data point. Specifically, a modified version of the multivariate Gaussian KDE is used to ensure that the estimated pdf has zero mean and identity covariance, consistent with the normalized properties of the data set obtained through PCA. The pdf estimate $p_{\mathbf{H}} : \mathbb{R}^\nu \to \mathbb{R}_+$ is given by:

\begin{align}\label{eq:pdf}
p_{\mathbf{H}}(\boldsymbol{\eta}) &= \frac{1}{N} \sum_{j=1}^N \pi_{\nu, \hat{s}_\nu}\left( \frac{\hat{s}_\nu}{s_\nu} (\boldsymbol{\eta}^{d,j} - \boldsymbol{\eta}) \right)
\end{align}

\noindent
\noindent
in which $\pi_{\nu, \hat{s}_\nu}$ is the positive function from $\mathbb{R}^\nu$ into $\, ]0, +\infty[\,$ defined, for all $\boldsymbol{\eta}$ in $\mathbb{R}^\nu$, by
\begin{equation}
\pi_{\nu, \hat{s}_\nu}(\boldsymbol{\eta}) = \frac{1}{(2\pi \hat{s}_\nu^2)^{\nu/2}} \exp\left( -\frac{\|\boldsymbol{\boldsymbol{\eta}}\|^2}{2 \hat{s}_\nu^2} \right)
\end{equation}
with $\| \boldsymbol{\eta} \|^2 = \eta_1^2 + \dots + \eta_\nu^2$, and where the positive parameters $s_\nu$ and $\hat{s}_\nu$ are defined by
\begin{equation}
s_\nu = \left\{ \frac{4}{N(2 + \nu)} \right\}^{1/(\nu + 4)}
\end{equation}

\begin{equation}
\hat{s}_\nu = \frac{s_\nu}{\sqrt{s_\nu^2 + \frac{N - 1}{N}}}
\end{equation}

\noindent
where the bandwidths $s_\nu$ and $\hat{s}_\nu$ control the width of the Gaussian kernels and hence the smoothness of the estimated density.  It can easily be verified that
\begin{equation}
\int_{\mathbb{R}^\nu} \boldsymbol{\eta} \, p_\textbf{H}(\boldsymbol{\eta}) \, d\boldsymbol{\eta} = \frac{\hat{s}_\nu}{s_\nu} \, \mathbf{m}' = \mathbf{0},
\tag{9}
\end{equation}

\begin{equation}
\int_{\mathbb{R}^\nu} \boldsymbol{\eta} \, \boldsymbol{\eta}^{\top} \, p_\textbf{H}(\boldsymbol{\eta}) \, d\boldsymbol{\eta} = 
\hat{s}_\nu^2 \left[ I_\nu \right] + \left( \frac{\hat{s}_\nu}{s_\nu} \right)^2 \frac{(N - 1)}{N} \left[ c' \right] = \left[ I_\nu \right].
\tag{10}
\end{equation}

This kernel density estimator $p_{\mathbf{H}}$ can now be used to define a product-form joint density over the random matrix $[\mathbf{H}] = [\eta^1, \dots, \eta^N]$, whose columns are assumed to be independent and identically distributed \cite{soize2016data} realizations of $\mathbf{H}$:

\begin{equation}
p_{[\mathbf{H}]}([\eta]) = p_{\mathbf{H}}(\boldsymbol{\eta}^1) \times p_{\mathbf{H}}(\boldsymbol{\eta}^2)\times \ldots \times p_{\mathbf{H}}(\boldsymbol{\eta}^N).
\end{equation}

\noindent
This formulation is essential for constructing the invariant measure in the stochastic differential equation-based sampling strategy introduced in the following sections \cite{soize2016data}. 
It encapsulates all the available statistical information extracted from the dataset and ensures that generated samples respect the observed empirical structure.

\subsection*{Construction of a full-order ISDE for Generating Realizations of Random Matrix $\mathbf{H}$}
\label{sec:full_SDE}

\noindent
Having established a nonparametric estimate for the pdf \( p_{[\mathbf{H}]} \), PLoM  leverages a stochastic construct capable of generating new samples that are statistically consistent with the given dataset and the estimated distribution. Specifically,  an Itô stochastic differential equation (ISDE) that admits the estimated joint density \( p_{\mathbf{[H]}} \) \textit{as its invariant measure} is formulated. The ISDE is defined on the space of \(\nu \times N\) real matrices and generates a diffusion process \(\{[\mathbf{U}(r)], r \geq 0\}\), where each evaluation of \([\mathbf{U}(r)]\) is an independent sample of the target random matrix \([\mathbf{H}] \in \mathbb{M}_{\nu,N}\). This process is coupled with a velocity process \([\mathbf{V}(r)]\), forming a second-order system analogous to a dissipative Hamiltonian dynamical system. The governing ISDE system is given by:

\begin{align}
d[\mathbf{U}(r)] &= [\mathbf{V}(r)]\,dr, \\
d[\mathbf{V}(r)] &= [\mathcal{L}([\mathbf{U}(r)])]\,dr - \frac{1}{2}f_0[\mathbf{V}(r)]\,dr + \sqrt{f_0} \, d[\mathbf{W}(r)],
\end{align}

\noindent
where:
\begin{itemize}
    \item \( [\mathbf{U}(r)] \in \mathbb{M}_{\nu,N} \) represents the position process (realizations of \([\mathbf{H}]\)),
    \item \( [\mathbf{V}(r)] \in \mathbb{M}_{\nu,N} \) represents the velocity process,
    \item \( f_0 > 0 \) is a damping parameter that controls the rate of convergence toward the stationary regime,
    \item \( [\mathbf{W}(r)] \) is a matrix-valued Wiener process composed of \(N\) independent Brownian motions in \(\mathbb{R}^\nu\),
    \item and \( [\mathcal{L}([\mathbf{U}(r)])] \) is a force-like term derived from the gradient of a potential function.
\end{itemize}

\medskip
\noindent
The process is initialized at $[\mathbf{U}(0)] = [\eta_d], [\mathbf{V}(0)] = [\mathcal{N}]$, where  \( [\mathcal{N}] \in \mathbb{M}_{\nu,N} \) is a random Gaussian matrix with independent standard normal entries, representing initial velocities.  $[\mathcal{L}]$ is a force term defined by the gradient of the potential function $V(\textbf{u}^l) = -\log q(\textbf{u}^l)$ as:
    \begin{equation}
        [\mathcal{L}([u])]_{kl} = -\frac{\partial}{\partial u_k^l} \log \{q(\textbf{u}^l)\}, \quad \textbf{u}^l = (u_1^l, u_2^l \ldots, u_{\nu}^l), \quad [u] = [\textbf{u}^1, \ldots, \textbf{u}^N]
    \end{equation}
   where $\textbf{u}^l \rightarrow q(\textbf{u}^l)$  is the continuously differentiable function such that 
\begin{equation}
q(\textbf{u}^l) =\sum_{j=1}^N \exp\left\{-\frac{1}{2s_{\nu}^2}  \left( \frac{\hat{s}_\nu}{s_\nu} \boldsymbol{\eta}_d^j - \textbf{u}^l \right)\right\}
\end{equation}.
 This ISDE framework ensures that the long-term statistical distribution of the samples \( [\mathbf{U}(r)] \) aligns with the target density \( p_{[\mathbf{H}]} \), while the damping and diffusion terms enforce ergodicity and guarantee convergence to equilibrium. The inclusion of the velocity component provides a second-order dynamic evolution that facilitates efficient exploration of the sample space, especially in high-dimensional settings.

\subsection*{Manifold Learning using Diffusion Maps}
\label{sec:dmaps}
\noindent
To capture the intrinsic low-dimensional geometry underlying the high-dimensional dataset, PLoM utilizes Diffusion Maps \cite{coifman2006diffusion}—a nonlinear dimensionality reduction technique grounded in spectral graph theory and stochastic diffusion processes. Unlike linear methods such as PCA, diffusion maps exploit local similarities in the data to reveal a global manifold structure. Diffusion maps leverage a symmetric, positive semi-definite kernel function that quantifies similarity between data points in the reduced space:

\begin{equation}
k_\varepsilon(\boldsymbol{\eta}, \boldsymbol{\eta}') = \exp\left(-\frac{\|\boldsymbol{\eta} - \boldsymbol{\eta}'\|^2}{4\varepsilon}\right),
\end{equation}

\noindent
where \(\varepsilon > 0\) is a scale parameter controlling the locality of the kernel. Smaller values of \(\varepsilon\) emphasize finer local structures, while larger values smoothen the geometry. Different metrics $||\cdot||$ are  possible (e.g., the $l^2$ norm).

Based on the kernel, the following matrices are constructed:

\begin{align}
[K]_{ij} &= k_\varepsilon(\boldsymbol{\eta}^{d,i}, \boldsymbol{\eta}^{d,j}), \\
[B] & =\quad[b]_{ii} = \sum_{j=1}^N [K]_{ij}, \\
[\tilde{K}] &= [B]^{-1}[K] [B]^{-1} \\
[D] &=\quad[d]_{ii} = \sum_{j=1}^N [\tilde{K}]_{ij}, \\
[\mathbb{P}] &= [D]^{-1}[\tilde{K}], \\
[\mathbb{P}_S] &= [D]^{1/2} [\mathbb{P}] [D]^{-1/2} = [D]^{-1/2} [\tilde{K}] [D]^{-1/2}.
\end{align}

\noindent
where, \([B]\) and \([D]\) are diagonal matrices of row-sums of \([K]\) and \([\tilde{K}]\) (normalized $[K]$), respectively, and \([\mathbb{P}]\) is a row-stochastic matrix representing transition probabilities of a discrete-time Markov chain. The symmetrized matrix \([\mathbb{P}_S]\) shares the same eigenvalues as \([\mathbb{P}]\), but is symmetric and more convenient for numerical diagonalization. The solution of the eigenvalue problem:

\begin{equation}
[\mathbb{P}_S] \boldsymbol{\varphi}_\alpha = \lambda_\alpha \boldsymbol{\varphi}_\alpha,
\end{equation}

\noindent
where \(\boldsymbol{\varphi}_\alpha \in \mathbb{R}^m\) (\(m\) is the number of retained diffusion modes) is a set of eigenvectors that capture orthogonal modes of diffusion along the manifold and \(\lambda_\alpha \in [0,1]\) are the corresponding eigenvalues, ordered such that \( \lambda_1 \geq \lambda_2 \geq \cdots \geq \lambda_m \).  A family of scaled eigenvectors is then constructed, which form the diffusion-maps basis. For a given analysis scale \(\kappa \in \mathbb{N}\), the diffusion coordinates are defined as:

\begin{equation}
\textbf{g}_\alpha = \lambda_\alpha^\kappa  [b]^{-1/2} \boldsymbol{\varphi}_\alpha, \quad \alpha = 1, \dots, m.
\end{equation}

\noindent
To calculate $m$, for an adapted value of $\epsilon$, one assumes that there exists a rapid decay of
the eigenvalues of $[\mathbb{P}]$ beyond a few leading ones. This embedding is actually a localization of the graph to the $m$-dimensional dominant eigenspace of $[\mathbb{P}]$.
These vectors \(\textbf{g}_\alpha \in \mathbb{R}^m\) constitute an orthogonal basis that reflects the manifold structure discovered by the diffusion process. Collecting them column-wise yields the matrix \([g] \in \mathbb{M}_{N,m}\), which serves as a reduced geometric embedding of the original dataset. 

One such challenge is the existence of ``repeated harmonic eigendirections'', which obscures the detection of the true dimensionality of the underlying manifold and arises when several embedding coordinates are harmonics of each other,  parametrizing the same direction in the intrinsic geometry of the data set.  To assess whether the $k$-th eigenvector $\phi_k$ corresponds to a new, unique eigendirection or is a repeated harmonic of the previous ones, the authors in \cite{dsilva2018parsimonious}  define a normalized leave-one-out cross-validation error based on local linear regression:

\begin{equation}
r_k = \sqrt{
\frac{
\sum_{i=1}^{n} \left( \phi_k(i) - \left( \hat{\alpha}_k(i) + \hat{\beta}_k(i)^T \Phi_{k-1}(i) \right) \right)^2
}{
\sum_{i=1}^{n} \phi_k(i)^2
}
}
\end{equation}
Here, $\Phi_{k-1}(i) = [\phi_1(i), \dots, \phi_{k-1}(i)]^T$ is the vector of the previous $(k-1)$ eigenvectors at point $i$, and the coefficients $\hat{\alpha}_k(i), \hat{\beta}_k(i)$ are obtained from locally weighted linear regression. A small $r_k$ suggests that $\phi_k$ is a harmonic (i.e., repeated eigendirection), while a large $r_k$ indicates a unique new direction.



\noindent

\subsection*{Reduced-Order ISDE}

\noindent
Once the data manifold has been uncovered using Diffusion Maps and the reduced-order representation $[\mathbf{H}] = [\mathbf{Z}] [g]^\top$ has been established, PLoM generates additional realizations that remain concentrated on the manifold, thereby preserving the geometric and statistical features of the dataset \cite{soize2016data}. 
By substituting the low-dimensional representation into the full ISDE system and implementing a Galerkin procedure to minimize the approximation error, we obtain the following reduced-order ISDE defined on $\mathbb{M}_{\nu,m}$:

\begin{align} \label{eq:red_ito}
d[\mathcal{Z}(r)] &= [\mathcal{Y}(r)]\,dr, \\
d[\mathcal{Y}(r)] &= [\mathcal{L}([\mathcal{Z}(r)]]dr - \frac{1}{2}f_0[\mathcal{Y}(r)]\,dr + \sqrt{f_0} \, [d\mathcal{W}(r)],
\end{align}
where $[a] = [g]([g]^\top[g])^{-1}$ and
where we have the following change of variables:
\begin{align}
    \textbf{U}(r)&=[\mathcal{Z}(r)][g]^\top \in \mathbb{M}_{\nu,m}\\
    \textbf{V}(r)&=[\mathcal{Y}(r)][g]^\top \in \mathbb{M}_{\nu,m}\\
    [d\mathcal{W}(r)] &= [d\textbf{W}(r)][a]\\ 
   [\mathcal{L}([\mathcal{Z}(r)] &= [\mathcal{L}([\mathcal{Z}(r)][g]^\top)][a].
\end{align}

\medskip
\noindent
The initial condition for this reduced-order simulation is obtained by projecting the original data and initial velocities onto the reduced basis:

\begin{equation}
[\mathcal{Z}(0)] = [\eta_d] [a], \quad [\mathcal{Y}(0)] = [\mathcal{N}] [a].
\end{equation}

\noindent
This setup ensures that the generated trajectories remain confined to the low-dimensional manifold, and that samples evolve under dynamics consistent with the estimated density. The reduced-order ISDE benefits from significantly lower computational cost and improved numerical stability due to the low dimensionality of the projected space. Moreover, since the diffusion basis was constructed to capture the intrinsic structure of the data, the samples generated in this reduced space are naturally well-aligned with the geometry of the original dataset. Finally, the generated samples are transformed back to the original data space $\mathbb{R}^n$ to obtain realizations of the random vector $\mathbf{X}$ that can be interpreted in the context of the application. Each simulated realization $[\eta_s]$ of the normalized random matrix $[\mathbf{H}]$ is recovered using the relation:

\begin{equation}
[\eta_s] = [\mathcal{Z}(l, \rho)] [g]^\top,
\end{equation}

\noindent
where $\rho= M_0\cdot r$ is chosen according to the ergodicity properties of the ISDE (typically after a sufficient relaxation time to ensure stationarity) and $l=1, \ldots, n_{\text{MC}}$. These samples are then mapped back to the original data space using the inverse of the PCA:

\begin{equation}
[x_s^l] = [\underline{x}] + \boldsymbol{\phi} [\mu]^{1/2} [\eta_s^l], \quad l=1, \ldots, n_{\text{MC}}
\end{equation}
For numerically solving the reduced-order ISDE,
the Störmer–Verlet scheme is usually utilized, since it preserves energy for non-dissipative Hamiltonian dynamical systems.

\section{Challenges in PLoM}
\label{sec:challenges}
\noindent
Since its introduction in 2016, the PLoM framework and its extensions \cite{soize:2024, soize2019entropy} have been extended to address increasingly complex small-data problems, including non-Gaussian Bayesian inference in high dimensions~\cite{soize2020sampling}, physics-informed learning from experimental data~\cite{soize2020physics},  PDE-constrained learning~\cite{soize2021probabilistic}, among others \cite{soize2024probabilistic, ghanem2018probabilistic}.  These advancements have enabled broader validation and refinement of the method. However, challenges remain when the dimension of the diffusion-map basis nears the number of samples, e.g., in extreme small-data settings and for highly complex densities. In such cases, PLoM may perform no better than standard MCMC methods that do not exploit manifold concentration. Moreover, the original PLoM formulation lacks of a systematic method for mapping points from the diffusion maps space back to the ambient data space. 

To address this, we propose a meaningfully enabling extension that integrates the recently developed Double Diffusion Maps \cite{evangelou2023double} with Geometric Harmonics (GH) \cite{coifman2006geometric} and solves a full-order ISDE in the reduced space rather than solving a reduced-order ISDE. By incorporating GH, we overcome the issue through a nonparametric reconstruction technique that enables smooth and accurate lifting from the reduced manifold back to the full ambient space, during the solution of the ISDE. Next, we briefly discuss GH and Double Diffusion Maps.

\section{Double Diffusion Maps}
\label{sec:ddmaps}

\noindent
The Double Diffusion Maps framework~\cite{evangelou2023double} builds upon the Diffusion Maps methodology (see Section~\ref{sec:dmaps}) to perform both dimensionality reduction and function extension. In this approach, a reduced set of non-harmonic coordinates, denoted by $\textbf{g}$, is first extracted from the leading non-harmonic eigenvectors. A corresponding basis, referred to as \textit{Latent Harmonics} and denoted by $\boldsymbol{\Psi}$, is then constructed using only these principal components. This latent basis provides a global interpolation mechanism, enabling scientific computations to be carried out directly within the reduced latent space. Although dimensionality is effectively reduced by retaining only the dominant non-harmonic eigenvectors—since the remaining ones are functionally dependent on these—this truncation limits the ability to extend general functions on the data manifold using GH, which typically rely on a complete basis. To address this limitation and recover the capacity for function extension while preserving the intrinsic low-dimensional structure of the manifold, the discarded eigenvectors are reconstructed through the computation of $\boldsymbol{\Psi}$ based on the reduced coordinates $\textbf{g}_{\alpha}$, where $\alpha=1,\ldots,m$. In this way, the Double Diffusion Maps approach restores the functional richness necessary for out-of-sample extension while maintaining computational efficiency in the reduced latent space.  For each coordinate in the set $\{\textbf{g}_{\alpha}^{(i)}\}_{i=1}^N$, we compute $$[K^\star]_{ij} = k^\star_{\epsilon}(\textbf{g}_{\alpha}^{(i)}, \textbf{g}_{\alpha}^{(j)}) = \exp\left(-\frac{||\textbf{g}_{\alpha}^{(i)} - \textbf{g}_{\alpha}^{(j)}||^2}{2\epsilon_2}\right)$$and calculate the first $m$ non-harmonic eigenvectors $\boldsymbol{\Psi} = \{\boldsymbol{\psi}_0, \dots, \boldsymbol{\psi}_{m -1}\}$, where $\boldsymbol{\psi}_i \in \mathbb{R}^N$, with corresponding eigenvalues $\boldsymbol{\sigma} = \{\boldsymbol{\sigma}_0, \dots, \boldsymbol{\sigma}_{m-1}\}$.  Given the function $h$ defined on the data, we project $h$ on these eigenvectors

\begin{equation}
h \rightarrow P_\delta h = \sum_{j=1}^m \langle h, \boldsymbol{\psi}_j \rangle \boldsymbol{\psi}_j
\label{eq:mapping}
\end{equation}
and we compute the Geometric Harmonics (GH) functions for $\boldsymbol{\phi}_{\text{new}}$ as:
\begin{equation}
\boldsymbol{\Psi}_j(\textbf{g}_{\alpha}^{\text{new}}) = \sigma_j^{-1} \sum_{i=1}^m k^\star_{\epsilon}(\textbf{g}_{\alpha}^{\text{new}}, \textbf{g}_{\alpha}^{(i)}) \psi_j(\textbf{g}_{\alpha}^{(i)})
\label{eq:psi_new}
\end{equation}
where $\psi_j(\textbf{g}_{\alpha}^{(i)})$ is the $i$-th component of the $j$-th eigenvector.
Finally, we estimate the value of the function for $\textbf{g}_{\alpha}^{\text{new}}$ as:
    \begin{equation}
    (Eh)(\textbf{g}_{\alpha}^{\text{new}}) = \sum_{j=1}^m \langle h, \boldsymbol{\psi}_j \rangle \boldsymbol{\Psi}_j(\textbf{g}_{\alpha}^{\text{new}})
    \label{eq:func}.
    \end{equation}

Algorithm \ref{alg:dim_reduction} depicts the basic steps for performing Double Diffusion Maps.


\begin{algorithm}[!htb]
\caption{Double Diffusion Maps}\label{alg:dim_reduction}

\begin{algorithmic}[1]
\Require A dataset $\textbf{X}=\{\textbf{x}_i\}_{i=1}^N \subset \mathbb{R}^n \sim p_{\textbf{x}(\cdot)}$
\State \textbf{Diffusion Maps} for Dimensionality Reduction: See Section \ref{sec:dmaps}

\State \textbf{Double Diffusion Maps}:
\begin{itemize}
\item For each $\textbf{g}_{\alpha}$ with $\alpha=1,\ldots,m$, compute  the kernel $k^\star_{\epsilon}(\textbf{g}_{\alpha}^{(i)}, \textbf{g}_{\alpha}^{(j)})$.
\item  Compute the $l$ first eigenvectors $\boldsymbol{\Psi} = \{ \boldsymbol{\psi}_0, \ldots, \boldsymbol{\psi}_{l-1} \}$ of $k^\star_{\epsilon}$, where $\boldsymbol{\psi}_i \in \mathbb{R}^N$.
\item Learn the mapping: $\mathsf{h} \to P_\delta \mathsf{h}$ using Eq.~(\ref{eq:mapping}).
\end{itemize}

\State \textbf{Latent Harmonics}:
\begin{itemize}
\State Compute the GH functions for $\boldsymbol{\Psi}_j(\textbf{g}_{\alpha}^{\text{new}})$  using Eq.~(\ref{eq:psi_new})
\item Estimate the value  $(E\mathsf{h})(\textbf{g}_{\alpha}^{\text{new}})$ with Eq.~(\ref{eq:func}).
\end{itemize}
\end{algorithmic}
\end{algorithm}

\subsection{Geometric Harmonics for out-of-sample extension}

\noindent
The GH framework, originally introduced by Coifman and Lafon~\cite{coifman2006geometric}, provides a powerful method for extending empirical functions defined on a dataset to new, unseen data points. This framework builds upon the Nyström extension  and is especially suited for data that lie on nonlinear manifolds embedded in high-dimensional spaces. Let $\textbf{X} = \{\textbf{x}_1, \dots, \textbf{x}_N\}$ be a finite dataset of interest in a high-dimensional ambient space, and let $f: \textbf{X} \to \mathbb{R}$ be a function defined only on the sample points. The objective is to estimate the values of $f$ at new, out-of-sample points $\textbf{x}_{\text{new}} \notin \textbf{X}$ in a way that is consistent with the geometry of the underlying manifold. GH constructs an affinity matrix $\textbf{W} \in \mathbb{R}^{N \times N}$ based on some kernel function, e.g., a Gaussian kernel:
\[
\textbf{W}_{ij} = \exp\left(-\frac{\|\textbf{x}_i - \textbf{x}_j\|^2}{\epsilon}\right),
\]
where $\epsilon$ is a scale parameter controlling the locality of the kernel. The matrix $\textbf{W}$ is symmetric and positive semi-definite. Performing eigendecomposition on $\textbf{W}$ yields a set of orthonormal eigenvectors $\{\boldsymbol{\psi}_\alpha\}_{\alpha=1}^N$ and corresponding non-negative eigenvalues $\{\sigma_\alpha\}_{\alpha=1}^N$ ordered such that $\sigma_1 \geq \sigma_2 \geq \cdots \geq \sigma_N \geq 0$. To ensure numerical stability and eliminate noisy components, only a subset of eigenvectors corresponding to eigenvalues above a threshold is retained. Specifically, the index set $S_\delta = \{\alpha : \sigma_\alpha \geq \delta \sigma_1\}$, where $0 < \delta < 1$ is defined. The function $f$ is then projected onto this reduced eigenspace:
\begin{equation}
P_\delta f = \sum_{\alpha \in S_\delta} \langle f, \boldsymbol{\psi}_\alpha \rangle \boldsymbol{\psi}_\alpha,
\end{equation}
where $\langle f, \boldsymbol{\psi}_\alpha \rangle$ denotes the standard inner product over the dataset $\textbf{X}$. To evaluate the function $f$ at a new point $\textbf{x}_{\text{new}} \notin \textbf{X}$, the basis functions $\boldsymbol{\psi}_\alpha$ is extended to the new point via the Nyström extension:
\begin{equation}
\boldsymbol{\Psi}_\alpha(\boldsymbol{\psi}_{\text{new}}) = \frac{1}{\sigma_\alpha} \sum_{i=1}^N \textbf{W}(\textbf{x}_{\text{new}}, \textbf{x}_i) \boldsymbol{\psi}_\alpha(\textbf{x}_i),
\end{equation}
where $\boldsymbol{\Psi}_\alpha(\boldsymbol{\psi}_{\text{new}})$ denotes the extension of the eigenfunction $\boldsymbol{\psi}_\alpha$. The extended value of the function $f$ at $\textbf{x}_{\text{new}}$ is then given as a linear combination of the extended basis functions:
\begin{equation}
(Ef)(\textbf{x}_{\text{new}}) = \sum_{\alpha \in S_\delta} \langle f, \boldsymbol{\psi}_\alpha \rangle \boldsymbol{\Psi}_\alpha(\textbf{x}_{\text{new}}).
\end{equation}
This formulation ensures that the interpolation remains smooth and respects the geometry captured by the kernel-defined spectral decomposition. As a results, GH effectively constructs an interpolant that captures both local and global variations in the function $f$ across the manifold where the data reside. One of the key strengths of the GH framework lies in its ability to perform nonlinear extension of functions, which allows it to capture complex, nonlinear trends in data far more effectively than traditional linear methods that may fail to represent curved or intricate structures embedded in high-dimensional spaces. This is due to the fact that GH leverage the spectral decomposition of a kernel-based affinity matrix to construct smooth extensions that naturally align with the geometry of the underlying manifold. Furthermore, the introduction of a spectral threshold parameter, denoted by $\delta$, enables selective filtering of eigencomponents based on their corresponding eigenvalues, offering precise control over the smoothness, regularity, and noise sensitivity of the resulting extension. This spectral filtering mechanism allows users to discard high-frequency components associated with small eigenvalues that may correspond to noise or insignificant variations in the data.  In addition, parameter $\delta$ helps control the ill-posedness of the scheme. As we retain more eigenvectors, we capture higher frequencies—but eventually, small eigenvalues begin to dominate. Since their magnitude approaches zero, this can cause significant numerical instability. Lastly, because the basis functions used in this approach are constructed directly from the data itself, they are inherently data-adaptive and capable of conforming to the shape, distribution, and complexity of the sampled manifold, making the method highly flexible and robust in real-world applications where the geometry is not known \textit{a priori}.

\section{Proposed method}
\label{sec:method}

\noindent
The present extension of the PLoM algorithm begins by applying Diffusion Maps to the high-dimensional dataset $[\mathbf{X}] \subset \mathbb{R}^n$, extracting the $m$ leading non-harmonic (latent) eigenvectors $\textbf{g} = [\textbf{g}_m]$ that parametrize the intrinsic manifold. Double Diffusion Maps  are then employed to learn the inverse mapping GH$_{-1}$: $[\textbf{g}_m] \rightarrow [\psi] \rightarrow [x_d]$, which reconstructs the high-dimensional structure from the diffusion coordinates. 
In this reduced space, the probability density function $p_{\boldsymbol{\Phi}}(\textbf{g})$ is estimated using a nonparametric approach (see Eq.~\ref{eq:pdf}) to model the empirical distribution of the data. Unlike the standard PLoM approach, which solves a reduced-order It\^o SDE, the proposed approach operates directly in the diffusion maps space  and solves a full-order It\^o SDE (see Section~\ref{sec:full_SDE}). Samples $[g_s^l]$ generated from this SDE are lifted to the ambient high-dimensional space using the GH$_{-1}$ map as $[x_s^l] = \text{GH}_{-1}([g_s^l])$. This process yields new realizations that are statistically consistent with the original dataset and remain concentrated on the learned manifold, enabling high-fidelity generative modeling and uncertainty propagation in complex systems. A step-by-step summary of the GH-PLoM algorithm is provided in Algorithm~\ref{alg:Ito}.


\begin{algorithm}[!htb]
\caption{GH-PLoM}\label{alg:Ito}

\begin{algorithmic}[1]
\Require A matrix version $[x_d]$ of the dataset $\textbf{X}$
\State \textbf{Diffusion Maps}:
\begin{itemize}
\item Find the matrix of leading non-harmonic diffusion maps  $[\textbf{g}_m]$ 
\item Perform Double diffusion maps to learn the inverse mapping: GH$_{-1}$: $[\textbf{g}_m] \rightarrow [\psi] \rightarrow [x_d]$
\end{itemize}




\State \textbf{Nonparametric statistical estimate of the density}
\begin{itemize}
\item Estimate the density $p_{\textbf{g}}(\textbf{g})$ using Eq.\ref{eq:pdf}: $p_{\textbf{g}}(\textbf{g}) = \frac{1}{N} \sum_{j=1}^N \pi_{\nu, \hat{s}_\nu}\left( \frac{\hat{s}_\nu}{s_\nu} (\textbf{g}^{m,j} - \textbf{g}) \right)$
\end{itemize}

\State \textbf{Nonlinear It\^o SDE}
\begin{itemize}

\item Solve the full-order It\^o SDE (see Section \ref{sec:full_SDE}) to generate samples $[g_s^l]$


\end{itemize}

\State \textbf{Lifting Operations}
\begin{itemize}
\item Lift to the ambient space using $[x_s] = \text{GH}_{-1}([g_s^l]), \quad l=1, \ldots, n_{\text{MC}}$
\end{itemize}

\end{algorithmic}
\end{algorithm}

\section{Numerical Examples}
\label{sec:examples}

\subsection{Example 1: Hermite Polynomial Dataset}
\label{sec:hermite}

\noindent
The first example is a synthetic toy dataset constructed using two-dimensional normalized \textit{Hermite basis functions}, commonly used in spectral methods for unbounded domains and as an orthogonal basis in Polynomial Chaos Expansions (PCE) for Gaussian uncertainties. Their smoothness, orthogonality, and fast decay also make them useful in quantum mechanics, image analysis, and reduced-order modeling.
The relevance of this example to us, however, is that $d$-dimensional Hermite polynomials are uncorrelated (i.e. orthogonal with respect to the Gaussian measure). Their lack of correlation prevents their reduction via correlation techniques (such as PCA or GP), in spite of their obvious $d$-dimensional latent structure. Let $x \in \mathbb{R}$. The univariate Hermite polynomials $\{h_n(x)\}$ are defined recursively by:
\begin{align}
    h_0(x) &= 1, \quad h_1(x) = x \nonumber \\
    h_{n+1}(x) &= x h_n(x) - n h_{n-1}(x), \quad n \geq 1. \nonumber
\end{align}

The associated normalized Hermite functions are given by:
\begin{equation}
    \psi_n(x) = \frac{h_n(x)}{\sqrt{n!}}. 
\end{equation}
In two dimensions, the multivariate Hermite functions are constructed via tensorization:
\begin{equation}
    \boldsymbol{\Psi}_{\boldsymbol{\alpha}}(\textbf{x}) = \psi_{\alpha_1}(x_1) \cdot \psi_{\alpha_2}(x_2), \quad \boldsymbol{\alpha} = (\alpha_1, \alpha_2) \in \mathbb{N}^2.
\end{equation}

We construct a dataset consisting of $N = 10,000$ samples of (selected) 2D Hermite polynomials of total degree $\leq 4$ (see Table \ref{tab:hermite_selected}). Each input $\textbf{x}^{(j)} = (x^{(j)}_1, x^{(j)}_2)$ is drawn independently from the bivariate standard normal distribution, i.e., $x^{(j)} \sim \mathcal{N}(\textbf{0}, \textbf{I}_2)$.  To simulate more realistic conditions we added noise to each Hermite polynomial sample to introduce a nontrivial stochastic component in the dataset  $\mathcal{D} =\{\textbf{y}_j\}_{j=1}^{N}$, where
\begin{equation}
    \mathbf{y}^{(j)} = \left[ \boldsymbol{\Psi}_{(0, 1)}(\textbf{x}^{(j)}), \ldots, \boldsymbol{\Psi}_{(2, 2)}(\textbf{x}^{(j)}) \right] \in \mathbb{R}^{9}
\end{equation}

\begin{table}[!htb]
\centering
\caption{2D Normalized Hermite polynomials $\boldsymbol{\Psi}_{\boldsymbol{\alpha}}(\mathbf{x})$ up to total order 4}
\begin{tabular}{cc}
\toprule
\textbf{Index $\boldsymbol{\alpha}$} & \textbf{Polynomial Expression} \\
\midrule
$(0,1)$   & $x_2$ \\
$(1,0)$   & $x_1$ \\
$(0,2)$   & $x_2^2 - 1$ \\
$(1,1)$   & $x_1 x_2$ \\
$(0,3)$   & $x_2^3 - 3x_2$ \\
$(1,2)$   & $x_1(x_2^2 - 1)$ \\
$(0,4)$   & $x_2^4 - 6x_2^2 + 3$ \\
$(1,3)$   & $x_1(x_2^3 - 3x_2)$ \\
$(2,2)$   & $(x_1^2 - 1)(x_2^2 - 1)$ \\
\bottomrule
\end{tabular}
\label{tab:hermite_selected}
\end{table}

\begin{figure}[htb!]
    \centering
    \includegraphics[width=0.8\textwidth, trim=20 20 20 0, clip]{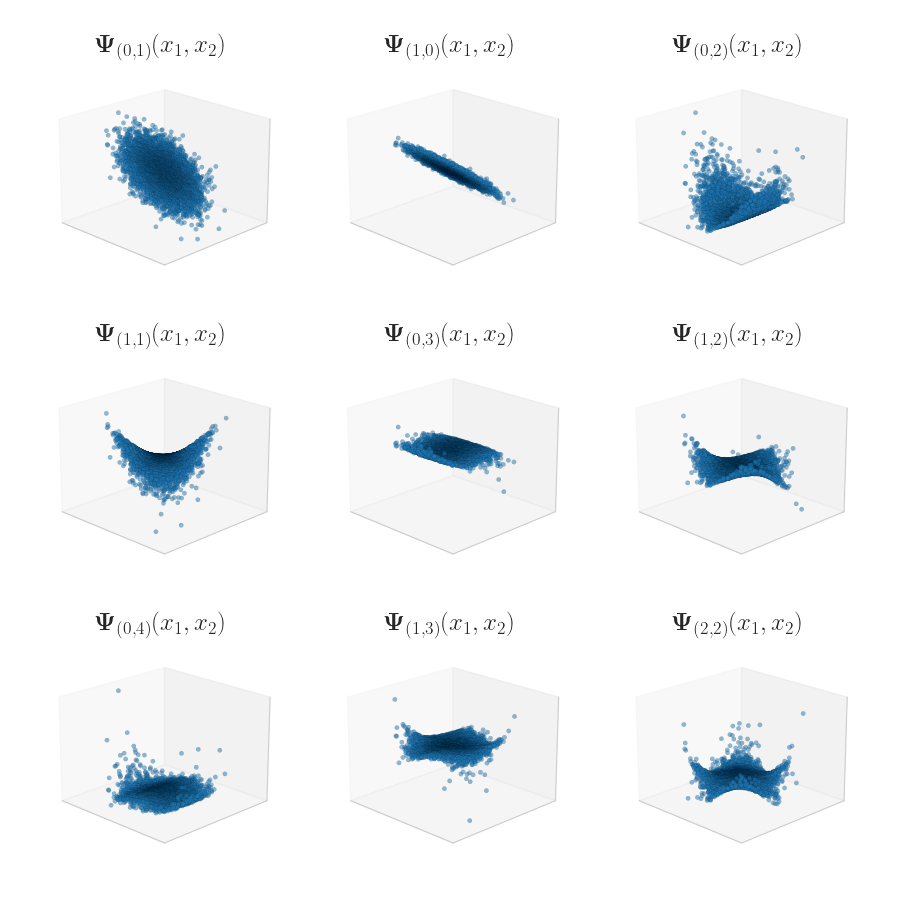}
    \caption{Visualization of the 2D Hermite polynomials samples.} 
    \label{fig:hermite_polys_samples}
\end{figure}

\begin{figure}[htb!]
    \centering
    \includegraphics[width=0.8\textwidth, trim=20 20 20 0, clip]{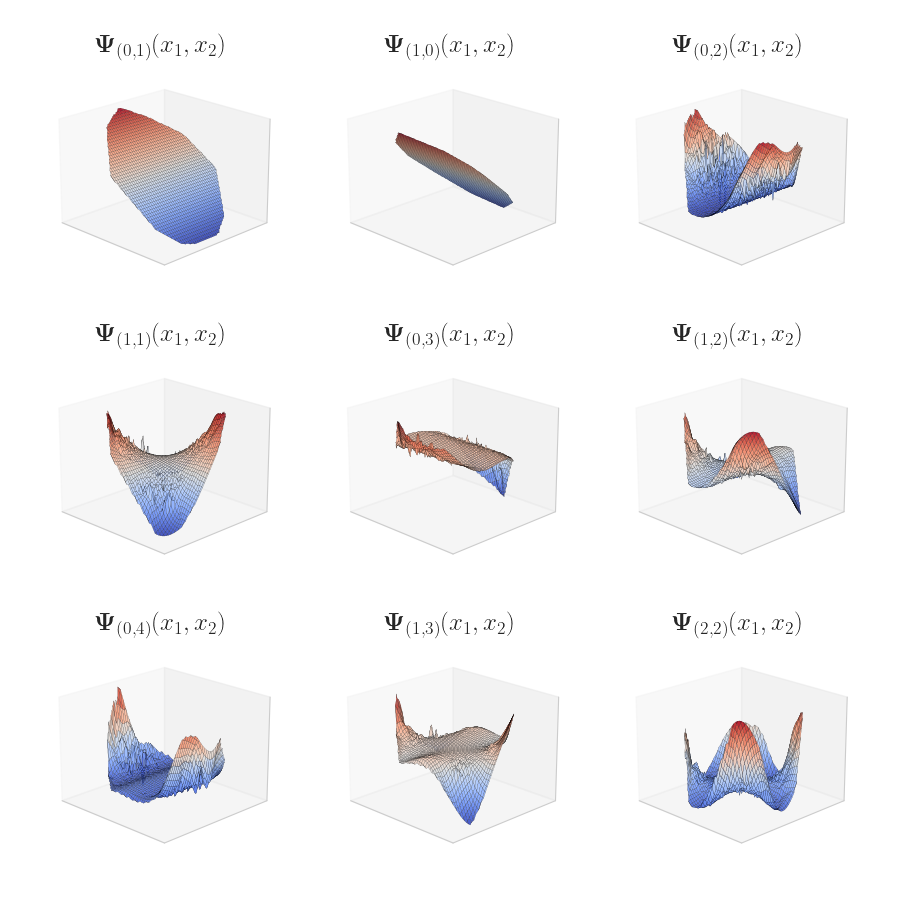}
    \caption{Visualization of selected 2D Hermite polynomials from Table~\ref{tab:hermite_selected} as surfaces. The color maps represent the values of $\textbf{H}_{\alpha}(\textbf{x})$.} 
    \label{fig:hermite_polys}
\end{figure}

Figure~\ref{fig:hermite_polys_samples} shows the Hermite polynomial samples showing the non-linear interactions between the input variables $x_1$ and $x_2$. Since each sample $\mathbf{x}^{(j)} = (x^{(j)}_1, x^{(j)}_2)$ is drawn independently from a bivariate standard normal distribution, capturing the variability of polynomial functions, especially near domain boundaries, can be challenging using the Gauss–Hermite basis, because only a few samples fall in regions with high variation, while most are concentrated near the origin. To address this non-uniform sampling, we apply \textit{min-max} scaling as a preprocessing step. This normalization maps the data to the range $[0, 1]$, helping to reduce sampling inconsistencies such as uneven point density across different regions. Figure~\ref{fig:hermite_polys} shows the corresponding 3D surface plots; The respective shapes of these surfaces highlight the increasing anisotropy and oscillatory behavior of Hermite polynomials as a function of their degree and contributes to the richness of the generated data manifold. For the construction of the Diffusion Maps embedding, we employed a Gaussian kernel with a bandwidth parameter set to 15 times the median of the pairwise distances. This increased scaling was necessary due to the sampling strategy, which required a broader kernel to adequately capture the underlying structure while maintaining locality. The affinity matrix is normalized using $\alpha=1$ to factor out the sampling density (as $N\rightarrow \infty$ the underlying operator converges to the Laplace Beltrami operator). The resulting Markov transition matrix is then spectrally decomposed to extract the dominant non-harmonic eigenvectors that capture the intrinsic geometry of the data.

\begin{table}[htb!]
\centering
\caption{Multivariate polynomials included in $\mathcal{D}_n$ for increasing $n$}
\begin{tabular}{cl}
\toprule
\textbf{$\mathcal{D}_n$} & \textbf{Basis Functions} \\
\midrule
$\mathcal{D}_0$ & $[\Psi_{(0,1)},  \Psi_{(0, 2)},  \Psi_{(0,3)}, \Psi_{(0, 4)}]$ \\
$\mathcal{D}_1$ & $[\Psi_{(0,1)}, \Psi_{(1,0)}, \Psi_{(0, 2)}]$ \\
$\mathcal{D}_2$ & $[\Psi_{(0,1)}, \Psi_{(1,0)}, \Psi_{0, 2)}, \Psi_{(1,1)}]$ \\
$\mathcal{D}_3$ & $[\Psi_{(0,1)}, \Psi_{(1,0)}, \Psi_{(0, 2)}, \Psi_{(1, 1)}, \Psi_{(0,3)}]$ \\
$\mathcal{D}_4$ & $[\Psi_{(0,1)}, \Psi_{(1,0)}, \Psi_{(0, 2)}, \Psi_{(1, 1)}, \Psi_{(0,3)}, \Psi_{(1,2)}]$ \\
$\mathcal{D}_5$ & $[\Psi_{(0,1)}, \Psi_{(1,0)}, \Psi_{(0, 2)}, \Psi_{(1,1)}, \Psi_{(0, 3)}, \Psi_{(1, 2)}, \Psi_{0, 4)}]$ \\
$\mathcal{D}_6$ & $[\Psi_{(0,1)}, \Psi_{(1,0)}, \Psi_{(0, 2)}, \Psi_{(1,1)}, \Psi_{(0,3)}, \Psi_{(1,2)}, \Psi_{(0, 4)}, \Psi_{(1,3)}]$ \\
$\mathcal{D}_7$ & $[\Psi_{(0,1)}, \Psi_{(1,0)}, \Psi_{(0, 2)}, \Psi_{(1,1)}, \Psi_{(0,3)}, \Psi_{(1,2)}, \Psi_{(0, 4)}, \Psi_{(1,3)}, \Psi_{(2, 2)}]$ \\
\bottomrule
\end{tabular}
\label{tab:hermite_dataset}
\end{table}

\begin{figure}[!htb]
    \centering
    \begin{subfigure}[b]{0.45\textwidth}
        \centering
        \includegraphics[width=\textwidth]{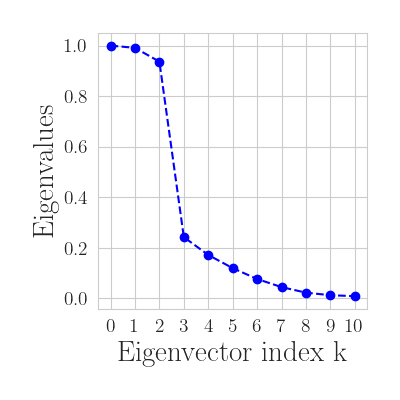}
        \caption{}
        \label{fig:residuals_d0}
    \end{subfigure}
    \hfill
    \begin{subfigure}[b]{0.45\textwidth}
        \centering
        \includegraphics[width=\textwidth]{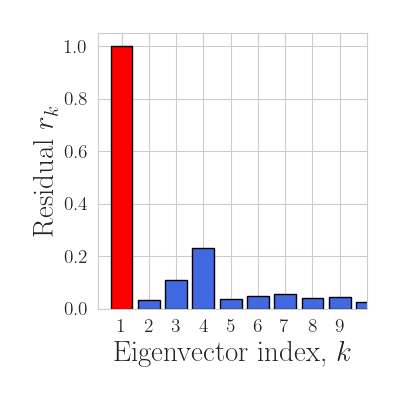}
        \caption{}
        \label{fig:residuals_d1}
    \end{subfigure}
    \caption{(a) Decay of eigenvalues and (b) normalized residuals $r_k$ from Diffusion Maps, for dataset $\mathcal{D}_0$. The red color indicates the selection of the eigenvectors that are non-harmonic. In this case, the intrinsic dimension of the dataset is one.}
    \label{fig:residuals_subplots}
\end{figure}

We examine different cases corresponding to increasingly complex data sets, that we label as $\mathcal{D}_n, \ n=1,\cdots, 8$. Table~\ref{tab:hermite_dataset} shows the specific multivariate polynomials included in each $\mathcal{D}_n$. The data sets $\mathcal{D}_n$, with $n=1,\ldots,7$ are constructed by gradually adding functions of the form $\Psi(i,j)$, enabling higher order polynomials and thus increasingly more complex interactions between $x_1$ and $x_2$. This hierarchical construction provides a flexible framework for analyzing datasets of varying complexity. We also include the data set \( \mathcal{D}_0 \), which includes bases that depend solely on \( x_2 \). This data set serves as a ``consistency check'' to verify that a single latent dimension underlies the data and that Diffusion Maps can effectively identify it. Figure~\ref{fig:residuals_subplots} shows (a) the eigenvalue spectrum and (b) the normalized residuals \( r_k \) calculated using the local linear regression method proposed by Dsilva et al.~\cite{dsilva2018parsimonious} for the data set \( \mathcal{D}_0 \). Eigenvectors identified as non-harmonic are highlighted in red. In this case, the intrinsic dimensionality of the dataset is one.

\begin{figure}[htb!]
    \centering
    \includegraphics[width=\textwidth]{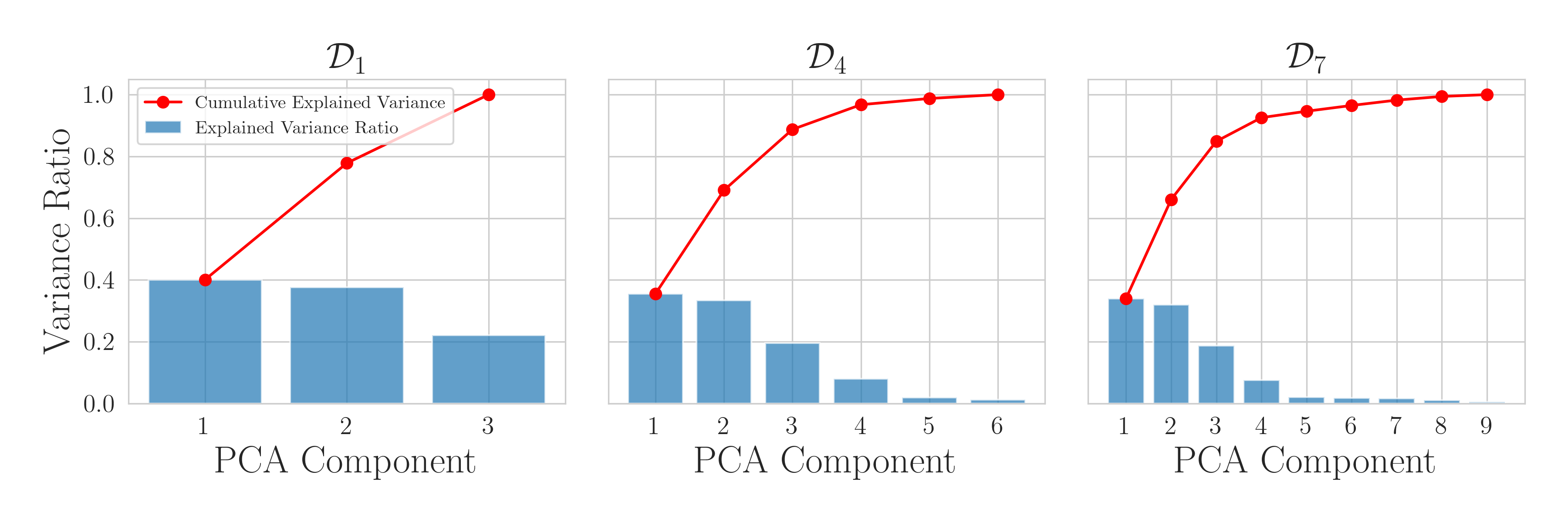}
    \caption{Spectrum of PCA eigenvalues for dataset $\mathcal{D}_1, \mathcal{D}_4$ and $\mathcal{D}_7$. Bars represent the explained variance ratio of individual principal components, while the red line shows the cumulative explained variance.}
    \label{fig:hermite_pca}
\end{figure}

 We note that the features of all the datasets $\mathcal{D}_n$ are orthogonal with respect to Gaussian measure (i.e. uncorrelated) modulo additive noise. This suggests that a PCA approach would not be able to reduce the dimension of the dataset. Figure~\ref{fig:hermite_pca} shows the explained variance ratio (bars) and cumulative explained variance (red line) for the first few principal components in six datasets $\mathcal{D}_1$ to $\mathcal{D}_7$. In each case, the PCA decomposes the data into orthogonal components ranked by their contribution to the total variance. As the dimensionality increases, the variance is distributed more evenly between components, with each contributing a smaller portion. 

\begin{figure}[htb!]
    \centering
    \includegraphics[width=\textwidth]{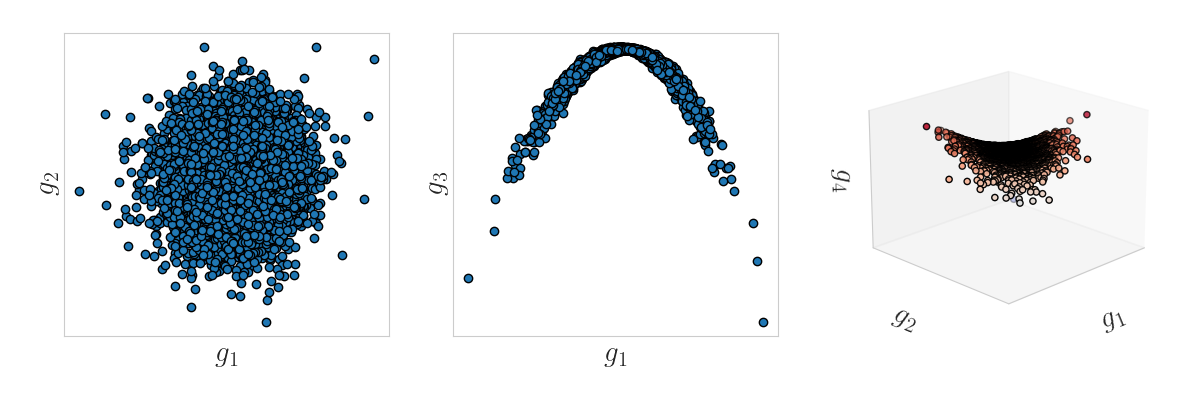}
    \caption{Visualization of the leading Diffusion Maps coordinates. 
    The left subplot shows the scatter plot of the first two non-harmonic eigenvectors $(\textbf{g}_1, \textbf{g}_2)$, which form the intrinsic coordinates of the underlying low-dimensional manifold. 
    The middle subplot displays the projection onto $(\textbf{g}_1, \textbf{g}_3)$, where $\textbf{g}_3$ is a harmonic extension and appears as a smooth function of the non-harmonic coordinate $\textbf{g}_1$. 
    The right subplot presents a 3D scatter plot of $(\textbf{g}_1, \textbf{g}_2, \textbf{g}_4)$, illustrating how higher-order coordinates behave as functions of the intrinsic variables. }
    \label{fig:DMAPS}
\end{figure}

Figure~\ref{fig:DMAPS} shows the first four Diffusion Maps coordinates for the data set $\mathcal{D}_7$.     The left subplot shows the scatter plot of the first two non-harmonic eigenvectors $(\textbf{g}_1, \textbf{g}_2)$, which form the intrinsic coordinates of the underlying low-dimensional manifold. 
The middle subplot displays the projection onto $(\textbf{g}_1, \textbf{g}_3)$, where $\textbf{g}_3$ is a harmonic extension and appears as a smooth function of the non-harmonic coordinate $\textbf{g}_1$. The right subplot presents a 3D scatter plot of $(\textbf{g}_1, \textbf{g}_2, \textbf{g}_4)$, illustrating how higher-order coordinates behave as functions of the intrinsic variables. This suggests that the intrinsic geometry of the dataset can be effectively represented in a reduced two-dimensional embedding, highlighting the ability of Diffusion Maps to uncover low-dimensional structure in high-dimensional data.

\begin{figure}[htb!]
    \centering
    \includegraphics[width=\textwidth, trim=0 10 10 10, clip]{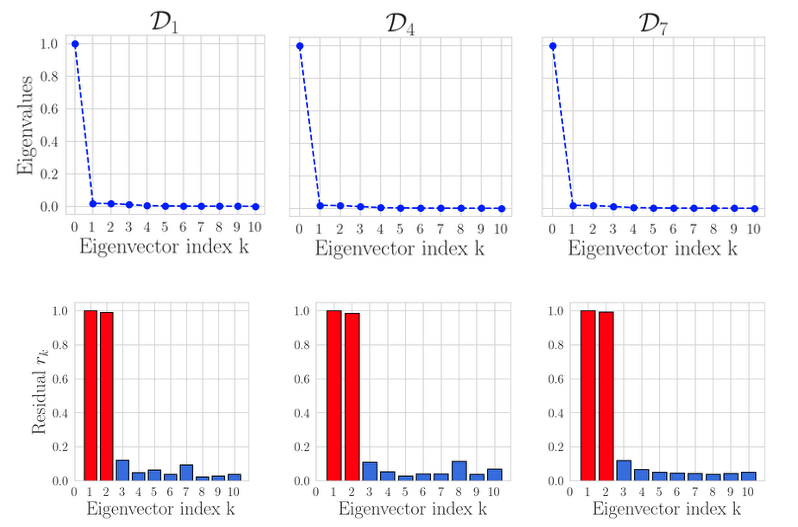}
    \caption{Top: Convergence of the eigenvalues in Diffusion Maps. Bottom: Normalized residuals $r_k$ from Diffusion Maps for each dataset in Table~\ref{tab:hermite_dataset}. The red color indicates the selection of the eigenvectors that are non-harmonic.}
    \label{fig:hermite_residuals_all}
\end{figure}


Figure~\ref{fig:hermite_residuals_all}~(Top) shows the convergence of the embedding eigenvalues of the diffusion maps, for each dataset $\mathcal{D}_1, \mathcal{D}_4$ and $\mathcal{D}_7$. 
For each eigenvector, we calculate the normalized residual $r_k$. The residuals with the largest magnitudes are shown in red in the bottom row of Figure~\ref{fig:hermite_residuals_all}, with the corresponding eigenvalues shown in the top row of the figure. We observe that in cases $\mathcal{D}_1$ to $\mathcal{D}_3$, two eigenvectors exhibit significantly higher residuals compared to the rest. These dominant eigenvectors capture the primary modes of variation in the dataset, indicating a low intrinsic dimensionality of the underlying manifold. 

\begin{figure}[htb!]
    \centering
    \includegraphics[width=\textwidth]{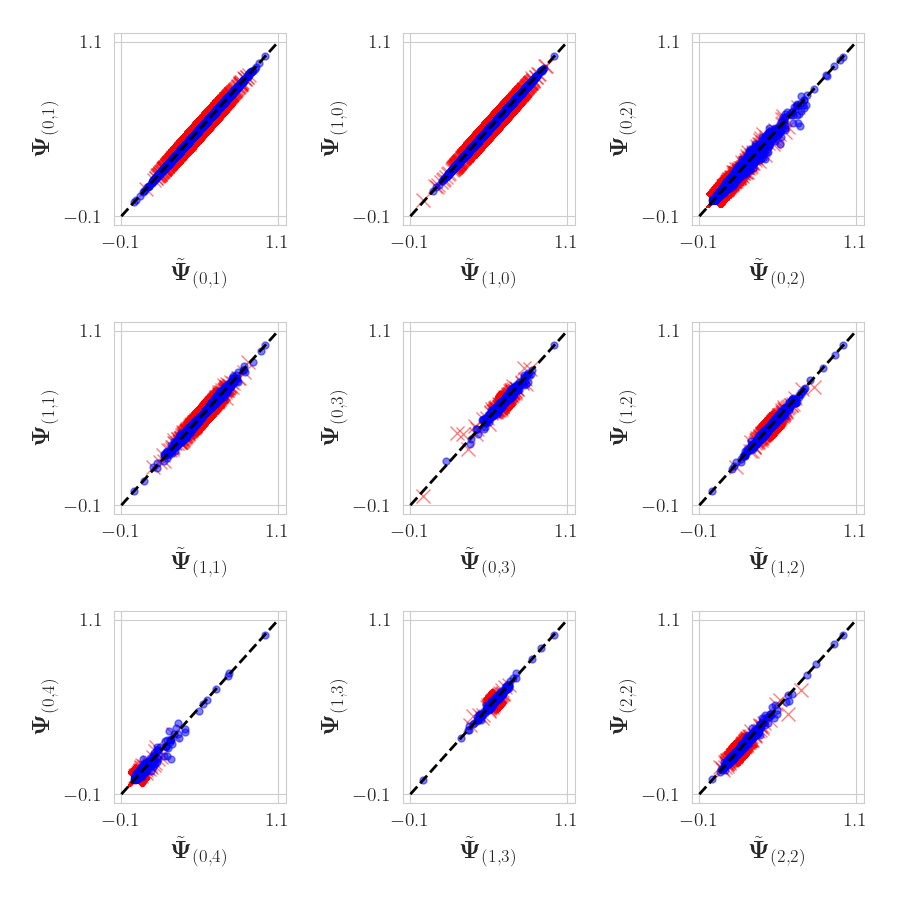}
\caption{The true values of the Hermite polynomials are plotted against the reconstructed by the GH for train (blue)
and test (red) points.}
    \label{fig:dmaps_GH}
\end{figure}

Figure~\ref{fig:dmaps_GH} presents the comparison between the exact values and the values lifted from the latent space basis functions using the GH for the data set $\mathcal{D}_{7}$. Each subplot shows a strong alignment between predicted and true values, with red points corresponding to the test data closely following the diagonal line. This demonstrates the accuracy and generalization ability of the GH model when reconstructing the selected functions from the latent representation.

%

\begin{figure}[htb!]
    \centering
    \includegraphics[width=\textwidth]{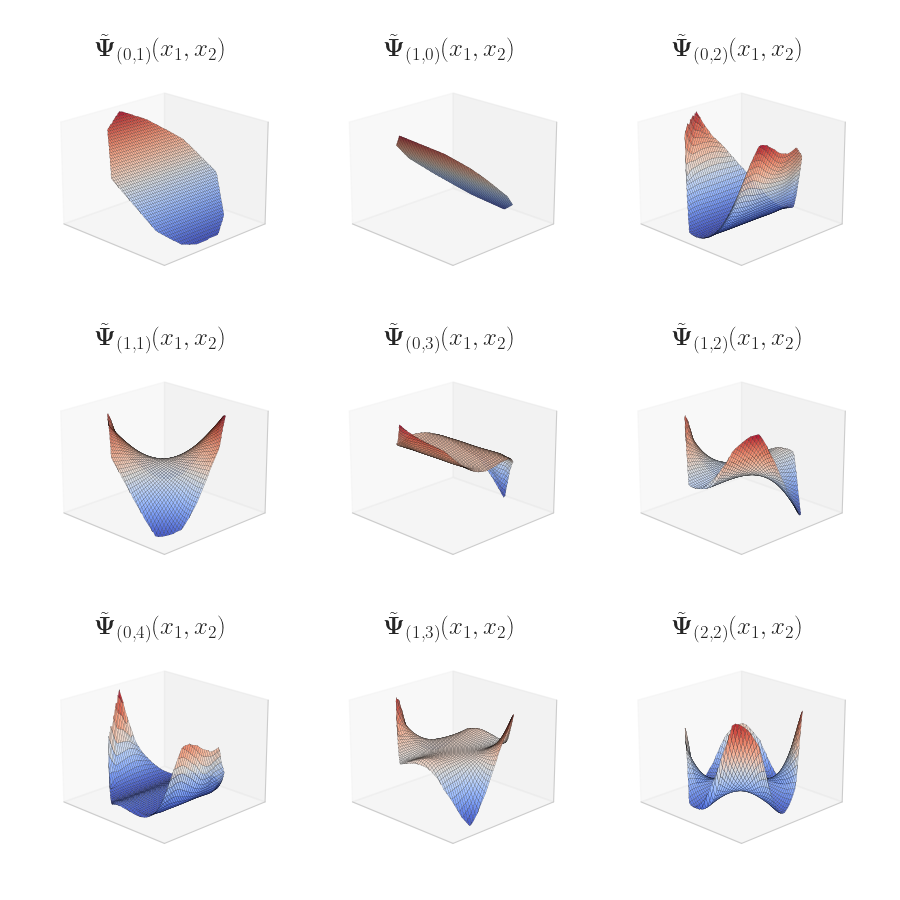}
    \caption{Expected values of the predicted Hermite basis conditioned on $x_1$ and $x_2$.}
    \label{fig:Plom_hermite}
\end{figure}

Finally, Figure~\ref{fig:Plom_hermite} illustrates the generated Hermite polynomials; we can see that the global structure and dominant features of the exact polynomials, particularly for low-order modes, are preserved. More specifically, the results are obtained by conditioning on $x_1$ and $x_2$ and predicting the expected value of the remaining columns. The spatial profiles of the generated polynomials exhibit correct nodal patterns and symmetry, confirming that the underlying manifold geometry is well captured.   Moreover, we can see that the noise has vanished since we are taking the expectation over 100 generated with PLoM realizations of the dataset. Overall, the generated basis functions are in good agreement with the analytical forms, which validates the ability of the proposed approach to approximate orthogonal polynomial structures in high-dimensional settings.

\subsection{Example 2: Flow Dynamics in Rotating
Detonation Engines}

\begin{figure}[htb!]
    \centering
    \includegraphics[width=0.5\textwidth]{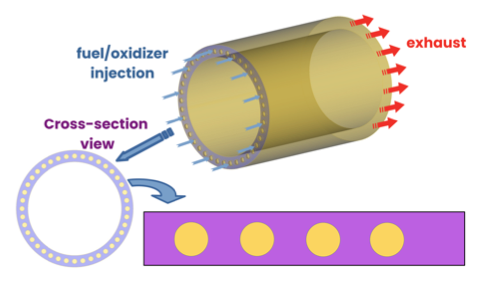}
\caption{The schematic diagram of inlets in rotating detonation combuster [REF].}
    \label{fig:model_wave}
\end{figure}

\noindent
In the second example, we test our methodology on a fairly complex mathematical physics problem constructed with a low-dimensional latent structure that evolves through complex multiphysics and multiscale interactions. Here again, we are able to disentangle the "essential dimensionality" and leverage it to carry out accurate yet efficient generative sampling.  Specifically,
we apply our proposed method to synthesize new configurations consistent with observations (simulation data) from a simplified model of a Rotating Detonation Engine (RDE), where the governing dynamics are described by compressible reacting flow equations coupled with detailed chemical kinetics. The domain of interest to us within an RDE is the annular space between two concentric cylinders, where fuel is injected axially through one section and combustion products exit through another. Detonation waves propagate azimuthally along this annular path. In our 2D simulations (see Figure~\ref{fig:model_wave}), the axial and azimuthal directions are treated as vertical and horizontal axes, respectively, while the radial direction is neglected due to its comparatively small extent. Although this thin radial region can influence detonation dynamics through strong wave reflections, its effects are considered secondary in the present model. Further, we consider a domain with only two fuel droplets that intercept a detonation wave. Details of the model can be found elsewhere \cite{kumar2025uncertainty}.

We construct a database of 897 synthetic solutions by varying the following four physical parameters that define the characteristics of the two fuel droplets: 
\begin{enumerate}
\item Oxygen volume fraction within the droplets;
\item Streamwise spacing between droplets; 
\item Vertical alignment angle of the droplet array;
\item Droplet diameter. 
\end{enumerate}
These parameters collectively influence the interplay between shock propagation, turbulent mixing, and localized energy release, thereby shaping the strength, structure, and stability of the detonation front—factors critical to RDE performance. Variations in inlet geometry, alignment, and fuel composition introduce uncertainty into the system, impacting the availability and distribution of reactants. 

The RDE configuration considered in this study is modeled as a two-dimensional annular domain with periodic azimuthal boundaries. The domain represents a simplified slice of the RDE chamber to capture essential detonation dynamics while reducing computational cost. The underlying physics are modeled using the compressible Navier-Stokes equations coupled with heat and species transport. Mathematical details of the model and its numerical implementation are detailed elsewhere \cite{kumar2025uncertainty}.

\begin{figure}[htb!]
    \centering
    \includegraphics[width=1.0\textwidth]{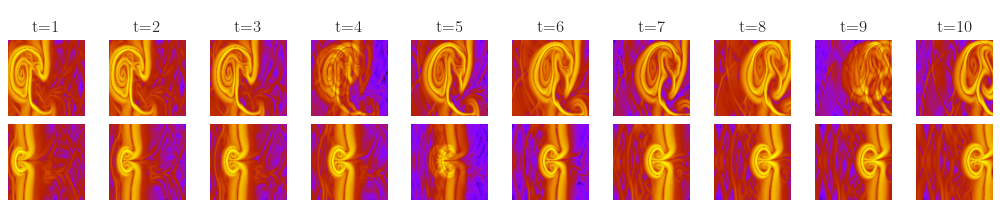}
    \caption{Two independent realizations of the wave propagation at different time steps $t=1$ to $t=10$.}
    \label{fig:realizations}
\end{figure}

\begin{figure}[htb!]
    \centering
    \includegraphics[width=1.0\textwidth]{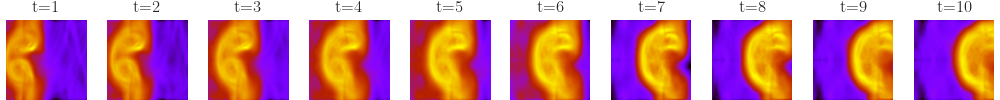}
    \caption{Temporal evolution of the mean wave field computed from 897 samples, illustrating the ensemble-averaged dynamics from $t=1$ to $t=10$.}
    \label{fig:mean_field}
\end{figure}

Each solution consists of 2D images capturing the propagation of the wave field over time. Figures~\ref{fig:realizations} and~\ref{fig:mean_field} illustrate the corresponding spatiotemporal dynamics. Figure~\ref{fig:realizations} displays two representative realizations, reflecting the variability introduced by differing fuel droplet injection and distribution patterns. These snapshots reveal complex structures, including vortices and interacting wavefronts, underscoring the system’s nonlinear behavior. In contrast, Figure~\ref{fig:mean_field} shows the mean wave field computed across all 897 samples, which smooths out random fluctuations and highlights the dominant propagation patterns.

\begin{figure}[htb!]
    \centering
    \includegraphics[width=\textwidth]{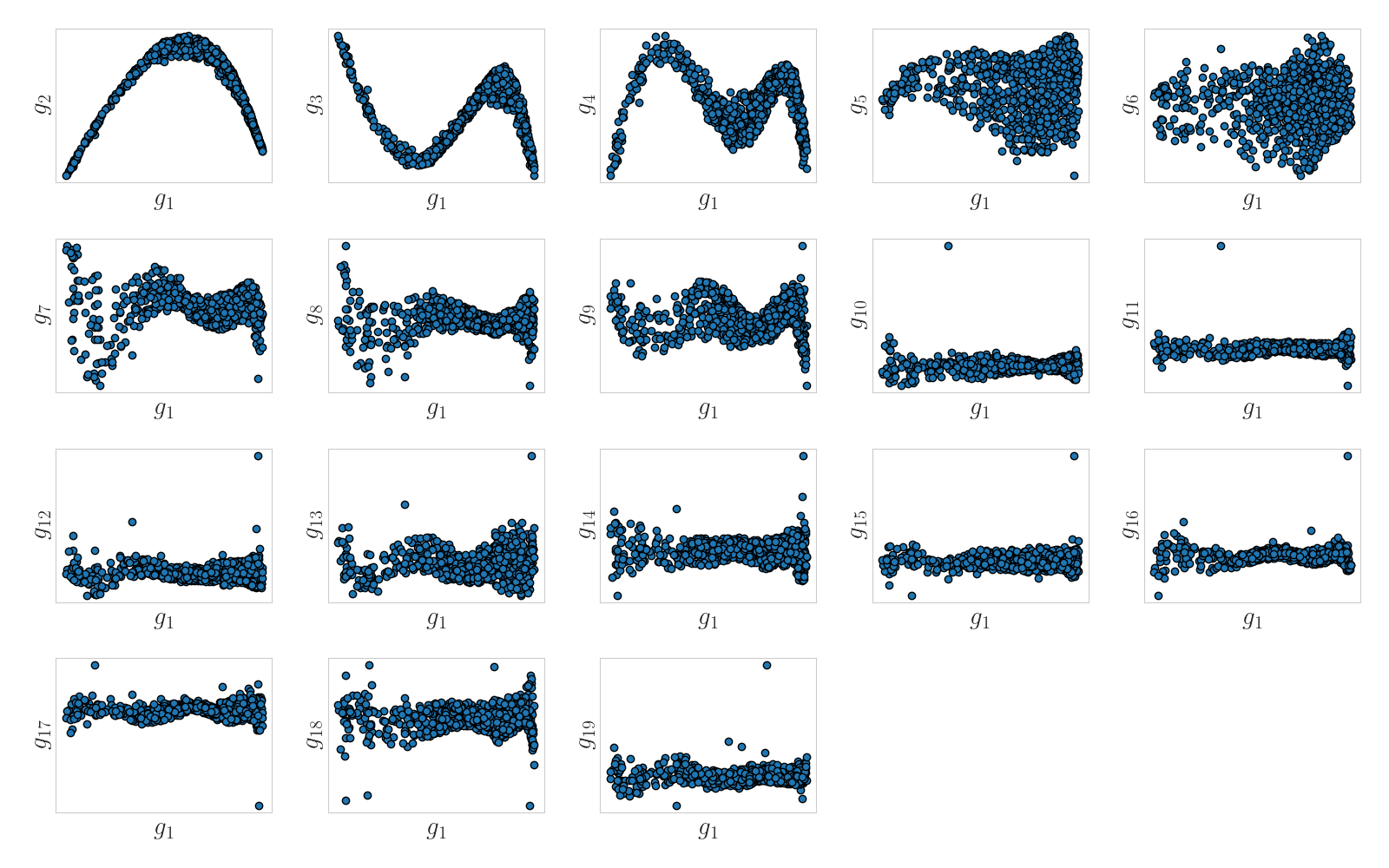}
    \caption{Diffusion Maps coordinates for the dataset as a function of $\textbf{g}_1$.}
    \label{fig:wave_dmaps}
\end{figure}

\begin{figure}[htb!]
    \centering
    \includegraphics[width=0.7\textwidth]{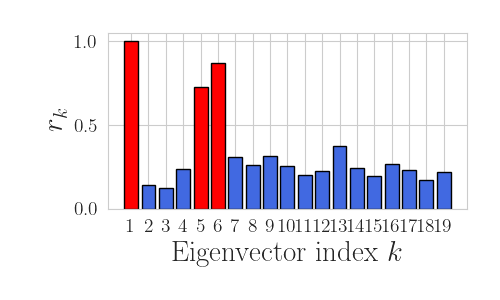}
    \caption{Normalized residuals $r_k$ obtained using the local linear regression method of Dsilva et al.~\cite{dsilva2018parsimonious}, computed for each eigenvector index $\lambda_k$ from the Diffusion Maps embedding of the wave propagation data.}
    \label{fig:residuals_wave}
\end{figure}

Figure~\ref{fig:wave_dmaps} illustrates the Diffusion Maps (we employed a Gaussian kernel with a bandwidth parameter set to 0.2 times the  median of the pairwise distances)
 coordinates plotted against the leading component $\mathbf{g}_1$. The plot highlights that the eigenvectors $\mathbf{g}_1$, $\mathbf{g}_5$, and $\mathbf{g}_6$ capture the dominant modes of variability and appear to be the most independent among the set. This observation indicates that the intrinsic structure of the dataset can be effectively embedded in a three-dimensional latent space. Further support for this reduced representation is provided by Figure~\ref{fig:residuals_wave}, which displays the normalized residuals $r_k$ computed using the local linear regression method proposed in~\cite{dsilva2018parsimonious}. These residuals, plotted against the corresponding eigenvalue indices $\lambda_k$, serve as a quantitative measure of the dimensionality of the manifold. Eigenvectors with the largest residuals are highlighted in red.  The residual analysis reinforces the hypothesis that a three-dimensional latent space is sufficient to describe the dataset's geometry. This low-dimensional structure can be interpreted physically: although the system is governed by four input parameters that can be independently varied, the underlying dynamics are primarily influenced by three emergent, non-dimensional factors—namely, the particle size, chemical composition, and spatial arrangement of the fuel. These factors collectively govern the evolution of the wavefront and resulting spatial features, giving rise to the observed manifold structure.

\begin{figure}[htb!]
    \centering
    \includegraphics[width=1.0\textwidth]{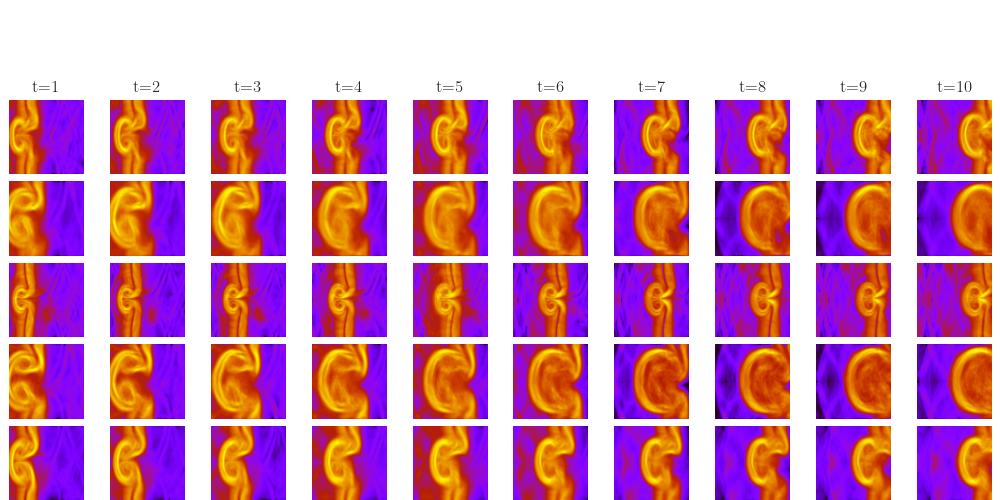}
\caption{Realizations of the generated samples using the proposed PLoM extension.}
    \label{fig:data_plom_wave}
\end{figure}

\begin{figure}[htb!]
    \centering
    \includegraphics[width=01.0\textwidth]{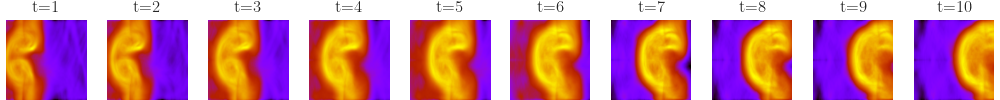}
\caption{
Temporal evolution of the mean wave field computed from the augmented samples, illustrating the ensemble-averaged dynamics from $t=1$ to $t=10$.}
    \label{fig:data_mean_plom_wave}
\end{figure}

Figure~\ref{fig:data_plom_wave} presents multiple generatively constructed wave fields are depicted for time steps $t=1$ through $t=10$, demonstrating the model's ability to capture both the variability and complexity of the underlying wave dynamics. Figure~\ref{fig:data_mean_plom_wave} shows the temporal evolution of the mean wave field, computed over 20{,}000 generated samples, effectively capturing the ensemble-averaged behavior of the system. For training and generation, each 2D wave snapshot was vectorized and the entire temporal sequence was concatenated into a single high-dimensional vector. The dimensionality of the vector used as input to the Diffusion Maps algorithm is $897 \times (10 \times 128 \times 128) = 897\times 163,840$.  This formulation allowed the model to learn the underlying spatiotemporal patterns and generate realistic trajectories that preserve the statistical properties of the original data.

\begin{figure}[htb!]
    \centering
    \includegraphics[width=01.0\textwidth]{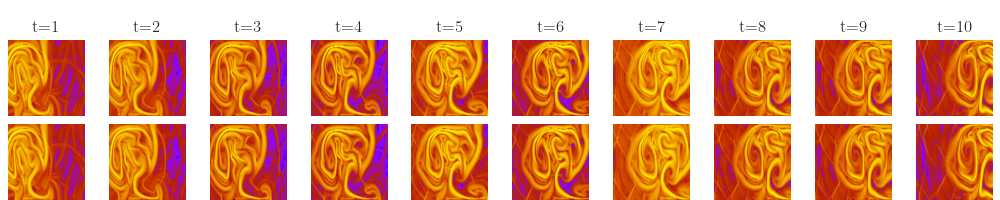}
\caption{Visualization of an extreme flow realization obtained by identifying the sample that is farthest from the rest in the reduced space. The corresponding physical-space field is shown for both the original training dataset (Top) and the augmented dataset (Bottom)}
    \label{fig:wave_extreme}
\end{figure}

Figure~\ref{fig:wave_extreme} illustrates an approach for visualizing extreme realizations of the flow field, rather than focusing solely on the mean behavior. Specifically, we identify the point that lies farthest from the rest in the reduced latent space. This outlier, which represents an extreme combustion scenario in the reduced representation, is then mapped back to the physical space to reveal the corresponding high-dimensional flow realization. By doing so, we gain insight into the most extreme behaviors present in the dataset, enabling a more comprehensive characterization of the system's variability beyond average trends.
A depiction of this extreme sample, in the original physical coordinates, is shown in Fig. (\ref{fig:wave_extreme}), computed both via the original training dataset (top) and the reconstructed image using our approach (bottom). This indicates a very accurate reproduction even for outlier samples.

\section{Conclusions}
\label{sec:conclusions}

\noindent
In this work, we introduced  generative learning framework for probabilistic sampling, based on a novel extension of the PLoM framework. While the original PLoM approach successfully leverages manifold geometry to generate statistically consistent samples, it faces challenges when the number of training samples is small and the diffusion maps basis becomes high-dimensional relative to the data size. To address this, we proposed, implemented, and demonstrated an enabling extension that usefully integrates the recently developed
Double Diffusion Maps with Geometric Harmonics (GH) and solves a full-order ISDE in the reduced space rather than solving a reduced-order ISDE. By incorporating GH, we overcome the issue through a nonparametric reconstruction technique that enables smooth and accurate lifting from the reduced manifold back to the full ambient space, during the
solution of the ISDE.  Furthermore, we addressed a key limitation of the original PLoM—namely, the absence of a principled {\em lifting mechanism} from the reduced space to the ambient space—by integrating Geometric Harmonics (GH), a smooth nonparametric interpolation method. This enabled us to reconstruct realistic high-dimensional realizations from the learned latent variables. The effectiveness of the proposed framework was demonstrated through two distinct applications: a synthetic dataset constructed using multivariate Hermite polynomials and a dataset from high-fidelity simulations a reactive detonation wave. These examples illustrate the ability of the method to capture complex dynamics, maintain statistical consistency, and generalize from limited observations. Future work will focus on scaling the approach to higher-dimensional problems and integrating additional physical constraints into the generative process.

\section*{Acknowledgments}
\noindent
The authors (DG and IGK) were partially supported by the Department of Energy (DOE) under Grant No. DE-SC0024162 and by the National Science Foundation (NSF) under Grant 2436738.  RG acknowledges support of an ONR MURI under grant No. N00014-22-1-2606. 

\bibliographystyle{unsrt}  
\bibliography{manuscript}  

\end{document}